\def\year{2020}\relax
\documentclass[letterpaper]{article} 
\usepackage{aaai20}  
\usepackage{times}  
\usepackage{helvet} 
\usepackage{courier}  
\usepackage[hyphens]{url}  
\usepackage{graphicx} 
\usepackage{graphicx}  
\usepackage{enumitem}
\usepackage{bbm}
\usepackage{multirow}
\usepackage{amsmath}
\usepackage{amssymb}
\usepackage{array}
\newcolumntype{P}[1]{>{\centering\arraybackslash}p{#1}}

\title{Learning 2D Temporal Adjacent Networks for \\ Moment Localization with Natural Language}

\author{Songyang Zhang \textsuperscript{\rm 1,2,}\thanks{Work performed as a research intern of Microsoft Research},
Houwen Peng\textsuperscript{\rm 2},
Jianlong Fu\textsuperscript{\rm 2},
Jiebo Luo\textsuperscript{\rm 1}\\
\textsuperscript{\rm 1}University of Rochester,
\textsuperscript{\rm 2}Microsoft Research\\
szhang83@ur.rochester.edu,
houwen.peng@microsoft.com,
jianf@microsoft.com,
jluo@cs.rochester.edu
}

\begin{document}

\maketitle

\begin{abstract}

We address the problem of retrieving a specific moment from an untrimmed video by a query sentence. This is a challenging problem because a target moment may take place in relations to other temporal moments in the untrimmed video. Existing methods cannot tackle this challenge well since they consider temporal moments individually and neglect the temporal dependencies. In this paper, we model the temporal relations between video moments by a two-dimensional map, where one dimension indicates the starting time of a moment and the other indicates the end time. This 2D temporal map can cover diverse video moments with different lengths, while representing their adjacent relations. Based on the 2D map, we propose a Temporal Adjacent Network (2D-TAN), a single-shot framework for moment localization. It is capable of encoding the adjacent temporal relation, while learning discriminative features for matching video moments with referring expressions. We evaluate the proposed 2D-TAN on three challenging benchmarks, i.e., Charades-STA, ActivityNet Captions, and TACoS, where our 2D-TAN outperforms the state-of-the-art.

\end{abstract}

\section{Introduction}

\begin{figure}[t!]
\centering
\includegraphics[width=0.475\textwidth]{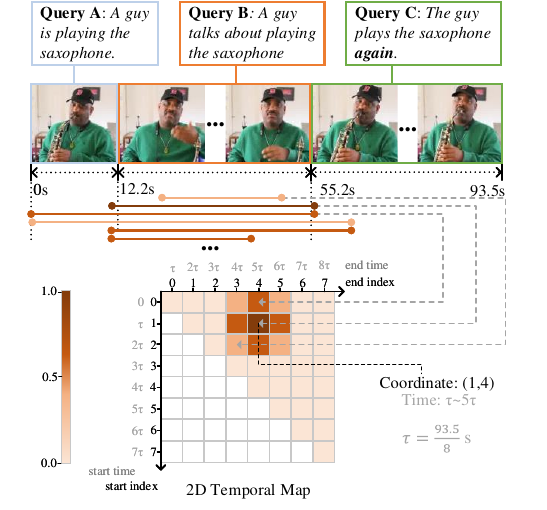}
\caption{Examples of localizing moments with natural language in an untrimmed video. 
In the two-dimensional temporal map, the \textit{black} vertical and horizontal axes represent the start and end frame indices while the corresponding \textit{gray} axes represent the corresponding start and end time in the video.
The values in the 2D map, highlighted by red color, indicate the matching scores between the moment candidates and the target moment.
Here, $\tau$ is a short duration determined by the video length and sampling rate.
}
\label{fig:task}
\end{figure}

Temporal localization is a fundamental problem of video understanding in computer vision.
Several related tasks are proposed for different scenarios, such as temporal action localization~\cite{SSN2017ICCV}, anomaly detection~\cite{hasan2016learning}, video summarization~\cite{song2015tvsum,chu2015video}, and moment localization with natural language~\cite{gao2017tall,hendricks17iccv}.
Among them, moment localization with natural language is the most challenging due to its flexibility and complexity of moment description.
This task is introduced recently by Gao \textit{et al.} and Hendricks \textit{et al.}~\cite{hendricks17iccv,gao2017tall}.
It aims to retrieve a temporary segment from an untrimmed video, as queried by a given natural language sentence.
For example, given a query ``\textit{a guy is playing the saxophone}'' and a paired video, the task is to localize the best matching moment described by the query, as shown in Figure.~\ref{fig:task}(Query A). Video moment localization with natural language has a wide range of applications, such as video question answering~\cite{lei2018tvqa}, video content retrieval~\cite{Shao_2018_ECCV}, as well as video storytelling~\cite{gella2018dataset}.

Most of the current language-queried moment localization models follow a two-step pipeline~\cite{gao2017tall,hendricks17iccv,Ge_2019_WACV,liu2018attentive,song2018val}.
Moment candidates are first selected from the input video with sliding windows. Each moment candidate is then matched with the query sentence to determine whether it is the target moment. 
This pipeline considers different moment candidates separately, thus neglecting their temporal dependencies. Therefore, it is difficult for current methods to model an moment that occurs in relation to other moments and predict the precise time boundary of the moment. For example, as shown in Figure~\ref{fig:task}(Query C), it targets to localize the query ``\textit{the guy plays the saxophone \textbf{again}}'' in the video. If the model only watches the temporal moments from the latter parts of the video, it cannot localize the described moment precisely.
Moreover, as shown in Figure~\ref{fig:task}(Query B), there are many temporal moments overlapping with the target moment (the visualized lines). 
These moments are related in visual content, but depict different semantics. It is difficult for previous methods to distinguish these visually similar moments since they process each moment candidate separately.

To address these problems, we propose a novel 2D Temporal Adjacent Networks (2D-TAN). The core idea is to localize the target moment on a two-dimensional temporal map, as presented in Figure~\ref{fig:task}.
Specifically, the $(i,j)$-th location on the 2D temporal map represents a candidate moment  from the time $i\tau$ to $(j+1)\tau$. 
This kind of 2D temporal map covers diverse video moments with different lengths, while representing their adjacent relations. 
In this fashion, 2D-TAN can perceive more moment context information when predicting whether a moment is related to other temporal segments. 
On the other hand, the adjacent moments in the map have content overlap but may depict different semantics. Considering them as a whole, 2D-TAN is able to learn discriminative features to distinguish them.

The main contributions of this paper are as follows.
\begin{itemize}[leftmargin=0.38cm]

\item{We introduce a novel \textit{two-dimensional temporal map} for modeling the temporal adjacent relations of video moments. Compared to previous methods, 2D temporal map enables the model to perceive more video context information and learn discriminative features to distinguish the moments with complex semantics.
}

\item{We propose a \emph{2D Temporal Adjacent Network}, i.e., 2D-TAN, for moment localization with natural language. 
Without any pre- or post-processing, 2D-TAN directly achieves competitive performance in comparison with the state-of-the-art methods on three benchmark datasets. 
\footnote{Our source code and model are available at \url{https://github.com/microsoft/2D-TAN}.}
}
\end{itemize}

\section{Related Work}

Temporal localization in untrimmed videos includes two major subfields: temporal action localization and moment localization with natural language. Temporal action localization aims to predict the start and end time and the label of the activity instance in untrimmed videos. The representative frameworks includes the two-stage temporal detection methods~\cite{SSN2017ICCV} and the one-stage single shot methods~\cite{lin2017single}. 
This task is limited to pre-defined simple actions and cannot handle complex activities in the real world. Therefore, moment localization with natural language~\cite{gao2017tall,hendricks17iccv} is introduced recently to tackle this problem.

Localizing moments in videos by referring expressions is a challenging task. It not only needs to understand video content, but also requires to align the semantics between video and language. 
For visual content understanding, several methods introduce local and global context in feature integration. 
Meanwhile, for video and language cross-modality alignment, existing methods exploit attention mechanism and RNN-based alignment. In the following, we discuss related methods from these two aspects.

\textit{Visual Content Understanding. } 
Context information is effective in visual content modeling. 
Existing methods integrate temporal context in two ways. 
One way is to use the whole video as the global context. Specifically, Hendricks \textit{et al}~\cite{hendricks17iccv} concatenate each moment feature with the global video feature~\cite{hendricks17iccv} as the moment representation.
Wang \textit{et al.} concatenate the semantic feature with the global video feature~\cite{wang2019language}. Another way is to use the surrounding clips as the local context for a moment.
Gao \textit{et al.}, Liu \textit{et al.}, Song \textit{et al.} and Ge \textit{et al.} concatenate the moment feature with clip features before and after the current clip as its representation~\cite{gao2017tall,liu2018attentive,song2018val,Ge_2019_WACV}.
 Since these methods model the context with a one-dimension sliding window, the moments longer than the window would be ignored. 
 Also, the long-range temporal dependencies across multiple windows would not be observed.
In contrast, our sampling strategy selects candidates from the entire input video, instead of a series of windows. This design enables segments with arbitrary lengths can be selected as candidates, which enables the model to perceive more context information and learn discriminative features.
Moreover, previous methods explore context information only on the visual feature, while ours models the context on the fused features of video and language. 

\textit{Video and Language Cross-Modality Alignment. } There are two methods for modeling video and language alignment: attention mechanism and sequential modeling.
For attention mechanism, the key idea is to align relevant visual features with the query text description by an attention module~\cite{vaswani2017attention}.
Hendricks \textit{et al.} and Zhang~\textit{et al.} apply a hard attention on moment features based on the sentence feature~\cite{hendricks18emnlp,zhang2019mm}, while
Liu \textit{et al.} and Xu \textit{et al.} use a soft attention~\cite{liu2018tmn,xu2019multilevel}.
Moreover, the visual-textual co-attention module is utilized to model the interaction between video and language ~\cite{liu2018crossmodal,song2018val,jiang2019cross,yuan2019to}. 
Instead of using the complex attention modules, our proposed 2D-TAN model only adopts a simple multiplication operation for visual and language feature fusion.

\begin{figure*}[!thb]
    \centering
    \includegraphics[width=0.95\textwidth]{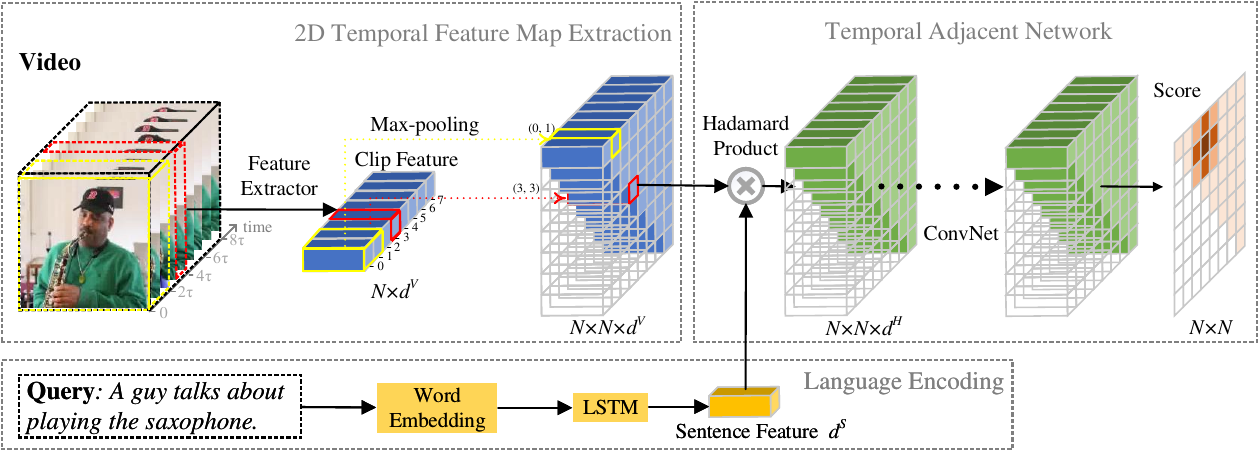}
    \caption{
    The framework of our proposed 2D Temporal Adjacent Network. It consists of a text encoder for language representation, a 2D temporal feature map extractor for video representation and a temporal adjacent network for moment localization.
    }
    \label{fig:pipeline}
\end{figure*}

For sequential modeling, the main idea is to align video with language by a recurrent neural network (RNN). 
The pioneering work is proposed by Chen \textit{et al.}, who 
design a recurrent module to temporally capture the evolving fine-grained frame-by-word interactions between video and sentence~\cite{chen2018temporally}. 
Zhang \textit{et al.} propose to apply a bidirectional GRU to align the features between video and language ~\cite{zhang2019cross}.
In contrast to these RNN-based methods where context information is gradually aggregated from clip representations, our proposed method explicitly models the context from moment representations via a 2D convolution network.

\section{Our Approach}
In this section, we first introduce the basic  formation of moment localization with natural language. Then, we propose the 2D Temporal Adjacent Network method. It consists of three steps: language representation, video representation, and moment localization. 
Figure~\ref{fig:pipeline} shows the framework of the proposed 2D-TAN approach.

\subsection{Problem Formulation} 
Given an untrimmed video $V$ and a sentence $S$ as a query, our task aims to retrieve the best matching temporary segment, i.e. the moment $M$, as specified by the query. More specifically, we denote the query sentence as $S=\{s_i\}_{i=0}^{l^S-1}$, where $s_i$ represents a word among the sentence, and ${l^S}$ is the total number of words. The input video stream is a frame sequence, i.e. $V=\{x_i\}_{i=0}^{l^V-1}$, where $x_i$ represents a frame in a video and $l^V$ is the total number of frames. 
The retrieved moment starting from frame $x_i$ to $x_j$ delivers the same semantic meaning as the input sentence $S$.

\subsection{Language Representation via Sequential Embedding}

We first extract the feature of an input query sentence. For each word $s_i$ in the input sentence $S$, we generate its embedding vector ${\bf{w}}_i \in \mathbb{R}^{d^S}$ by the GloVe word2vec model~\cite{pennington2014glove}, where $d^S$ is the vector length. Then, we sequentially feed the word embeddings $\{ {\bf{w}}_i \}_{i=0}^{l^S-1}$ into a three-layer LSTM network~\cite{hochreiter1997long}, and use its last hidden state as the feature representation of the input sentence, i.e. ${\bf{f}}^S\in \mathbb{R}^{d^{S}}$. The extracted feature encodes the language structure of the query sentence, thus describe the moment of interest.

\subsection{Video Representation via 2D Temporal Feature Map}
This section extracts the features of the input video stream, and encodes the features into a two-dimensional temporal feature map. 
For an input video stream, we first segment it into small video clips. 
Each video clip $v_i$ consists of $T$ frames. 
Then, we perform a fixed-interval sampling over the video clips, and obtain $N$ videos clips, denoted as $V=\{v_i\}_{i=0}^{N-1}$. 
For each sampled video clip, we extract its feature using a pre-trained CNN model (see Experiment section for details). To get a more compact representation, we pass the extracted feature through a fully-connected layer with $d^V$ output channels. The final representation of a sampled video clip is represented as ${\bf{f}}^V \in {\mathbb{R}}^{d^V}$, where $d^V$ is the feature dimension. 

The sampled $N$ video clips serve as the basic elements for moment candidate construction. Thus, we build up the feature map of moment candidates by the video clip features $ {{\{{\bf{f}}^V}\}_{i=0}^{N-1}}$.
Previous works extract moment features from clip features in two ways: pooling~\cite{hendricks17iccv} or stacked convolution~\cite{zhang2019man}.
In this work, we follow the pooling design. For each moment candidate, we max-pool the corresponding clip features across a specific time span, and obtain its feature  ${{\bf{f}}_{a,b}^M}=maxpool({{\bf{f}}_a^V}, {{\bf{f}}_{a+1}^V}, ..., {{\bf{f}}_b^V})$, where $a$ and $b$ represent the indexes of start and end video clips, and $0 \le a \le b \le N-1$,
Long-time moment candidates are pooled over serveral consecutive clips, while short-time ones are pooled over few clips. As a result, the features of moment candidates are contructed.
In addition, the alternative solution, i.e. stacked convolution, is also compared in our experiments.

\begin{figure}[t]
    \centering
    \includegraphics[width=0.475\textwidth]{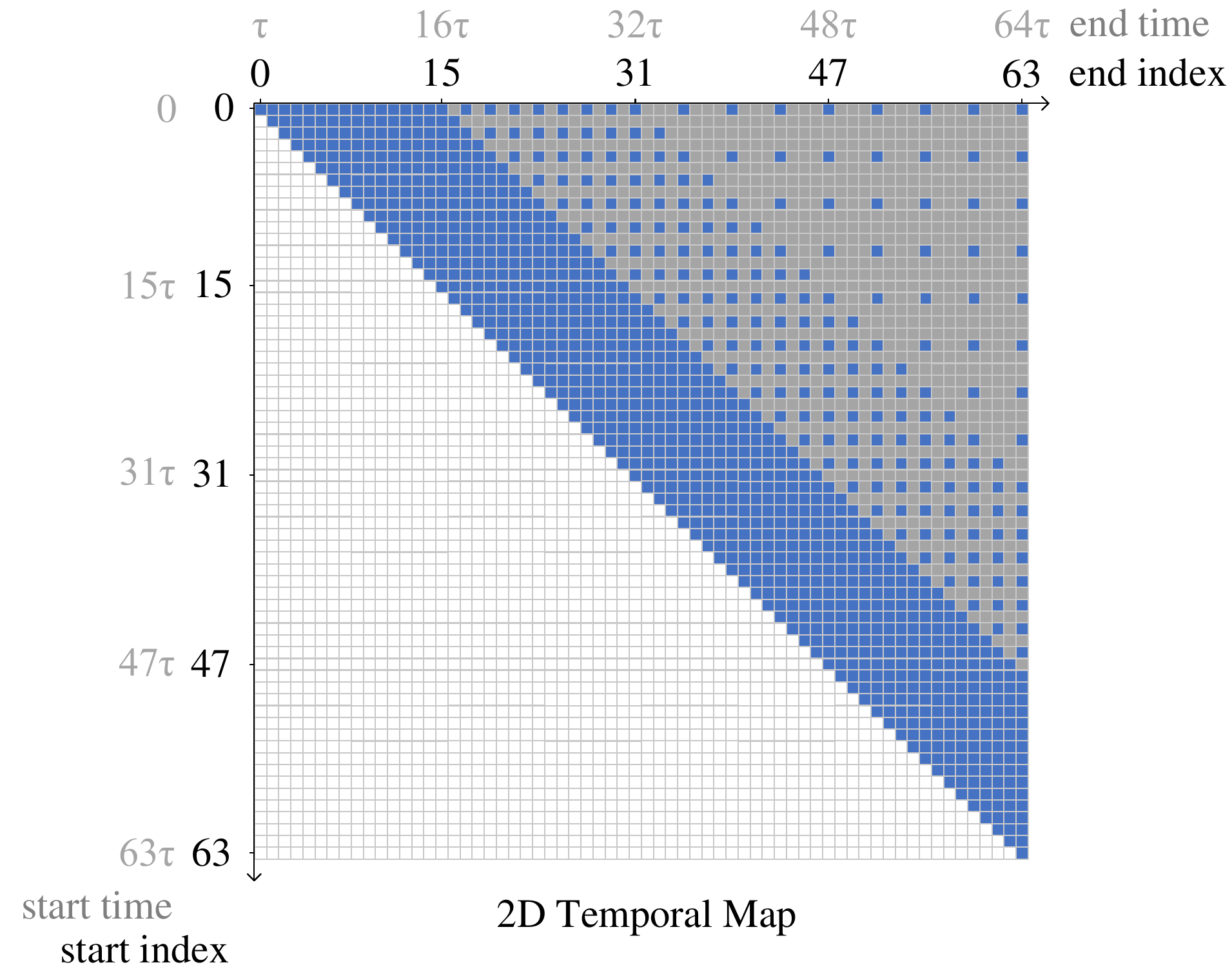}
    \caption{The selection of moment candidates when there are $N=64$ sampled clips in an untrimmed video. The upper triangular part of the 2D map enumerates all possible moment candidates starting from clip $v_a$ to $v_b$, while the lower triangular part is invalid.
    In our method, only the \textit{blue} points are selected as moment candidates.
    Best viewed in color.
    }
    \label{fig:candidate}
\end{figure}

Different from previous methods which directly operate on an individual video moment, we restructure the whole sampled moments to a 2D temporal feature map, denoted as $\mathbf{F}^M\in \mathbb{R}^{N\times N\times d^V}$. The 2D temporal feature map $\mathbf{F}^M$ consists of three dimensions: the first two dimensions $N$ represent the start and end clip indexes respectively, while the third one $d^V$ indicates the feature dimension.
The feature of a moment starting from clip $v_a$ to $v_b$ is located at $\mathbf{F}^M[a,b,:]$ on the feature map, where $\mathbf{F}^M[a,b,:] = {\bf{f}}_{a,b}^M$.
Noted that, the moment's start and end clip indexes $a$ and $b$ should satisfy $a\le b$.
Therefore, on the 2D temporal feature map, all the moment candidates locating at the region of $a>b$ are invalid, i.e. the lower triangular part of the map, as shown in Figure~\ref{fig:task}  and~\ref{fig:candidate}. The values in this region are padded with zeros in implementation.

The previous three steps introduce the feature extraction of moments, but do not specify how to select possible moments as candidates.
One simple way is to enumerate all the possible consecutive video clips as candidates. 
However, this strategy will bring much computational cost to subsequent moment-sentence matching and retrieval.
Therefore, we propose a sparse sampling strategy, as shown in Figure~\ref{fig:candidate}.
The key idea is to remove the  redundant moments which have large overlaps with the selected candidates.

Specifically, we densely sample moments of short duration, and gradually increase the sampling interval when the moment duration becomes long.
In more details, when the number of sampled clips is small, i.e. $N\le 16$, we enumerate all possible moments as candidates.
When $N$ becomes large, i.e. $N>16$, a moment starting from clip $v_a$ to $v_b$ is selected as the candidate when satisfying the following condition $G(a, b)$:
\begin{equation}
       G(a,b) \Leftarrow 
             (a~\emph{mod}~s {\tiny{=}} 0) ~~\&~~ ((b-s')~\emph{mod}~s {\small{=}} 0),
\label{eq:candidate1}
\end{equation}
where $a$ and $b$ are the indexes of clips, $s$ and $s'$ are defined as:
\begin{equation}
    \begin{aligned}
        s&=2^{k-1},\\
        s'&=
        \begin{cases}
        0 & \text{if $k=1$,}\\
        2^{k+2}-1 & \text{otherwise.}\\ 
        \end{cases}\\
    \end{aligned}
\label{eq:candidate2}
\end{equation}
Here, $k=\lceil\log_2(\frac{b-a+1}{8})\rceil$, and $\lceil \cdot \rceil$ is the ceil function.  
If $G(a,b)=1$, the moment is selected as the candidate, otherwise, it is not selected.
This sampling strategy can largely reduce the number of moment candidates, as well as the computational cost.

\subsection{Moment Localization via 2D Temporal Adjacent Network}
When both the language and video feature representations are ready, we predict the best matching moment queried by the sentence from all candidates. It mainly includes three continuous processes: multi-modal fusion, context modeling and score prediction.

We first fuse the 2D temporal feature map ${\bf{F}}^M$ with the encoded sentence feature ${\bf{f}}^S$. 
Specifically, we project these two cross-domain features into an unified subspace by fully-connected layers, and then fuse them through Hadamard product and $\ell_2$ normalization as

\begin{equation}
    \mathbf{F} = \| ({\bf{w}}^S \cdot {\mathbf{f}}^S \cdot {\mathbbm{1}}^T) \odot ({\bf{W}}^M \cdot {\mathbf{F}}^M) \|_F,
\label{eq:fusion}
\end{equation}
where ${\bf{w}}^S$ and ${\bf{W}}^M$ represents the learnt parameters of the fully connected layers, ${\mathbbm{1}}^T$ is the transpose of an all-ones vector, $\odot$ is Hadamard product, and $\|\cdot\|_F$ denotes Frobenius normalization.
We further build up the Temporal Adjacent Network over the fused 2D feature map $\bf{F}$.
The network architecture is simple, and only consists of $L$ convolutional layers with kernel size of $K$. The output of the network keeps the same shape as the input fused feature map through zero padding.
This design enables the model to gradually perceive more context of adjacent moment candidates, while learn the difference between moment candidates. Moreover, the receptive filed of the network is large, thus it can observe the whole content of whole video and sentence, resulting in learning the temporal dependencies.
It is worth noting that, within the 2D fused feature map, there are zero-padding regions. 
When performing convolutions over these regions, we only calculate the values on the valid location. In other words, zero-padding features are not taken into calculation.

Finally, we predict the matching scores of moment candidates with the given sentence on the 2D temporal map. The output feature of temporal adjacent network goes through a fully connected layer and a sigmoid function, then generates a 2D score map.
According to the candidate indicator $G(a,b)$ in Equation~(\ref{eq:candidate1}),
all the valid scores on the map are then collected, denoted as $P=\{p_i\}_{i=1}^C$, where $C$ is the total number of moment candidates.
Each value $p_i$ on the map represents the matching score between a moment candidate with the queried sentence. The  maximum value indicates the best matching moment.

\subsection{Loss Function}
During the training of our 2D-TAN, we adopt a scaled $IoU$ value as the supervision signal, rather than a hard binary score. 
Specifically, for each moment candidate, we compute its $IoU$ $o_i$ with the ground truth moment. The $IoU$ score $o_i$ is then scaled by two thresholds $t_{min}$ and $t_{max}$ as
\begin{equation}
\begin{aligned}
    y_i& =
    \begin{cases}
    0 & {{o_i} \le t_{min}}, \\
    \frac{o_i-t_{min}}{{t_{max}-t_{min}}} & {t_{min} < {o_i} <  t_{max}}, \\
    1 & {{o_i} \ge t_{max}}, 
    \end{cases}
\label{eq:iou}
\end{aligned}
\end{equation}
and $y_i$ serves as the supervision label. 
Our network is trained by a binary cross entropy loss as
\begin{equation}
    Loss =\frac{1}{C}\sum_{i=1}^C y_i \log p_i + (1-y_i) \log(1-p_i),
\label{eq:loss}
\end{equation}
where $p_i$ is the output score of a moment and $C$ is the total number of valid candidates.

\section{Experiment}
We evaluate the proposed 2D-TAN approach on three public large-scale datasets: Charades-STA~\cite{sigurdsson2016hollywood}, ActivityNet Captions~\cite{krishna2017dense} and TACoS~\cite{regneri2013grounding}.
In this section, we first introduce these datasets and our implementation details, and then compare the performance of 2D-TAN with other state-of-the-art approaches. Finally, we investigate the impact of different components via a set of ablation studies.

\subsection{Dataset}

\textit{~~~Charades-STA.} 
It contains $9,848$ videos of daily indoors activities. It is originally designed for action recognition and localization. Gao \textit{et al.}~\cite{gao2017tall} extend the temporal annotation (i.e. labeling the start and end time of moments) of this dataset with language descriptions and name it as Charades-STA. Charades-STA contains $12,408$ moment-sentence pairs in training set and $3,720$ pairs in testing set.

\textit{ActivityNet Captions.} It consists of $19,209$  videos, whose content are diverse and open.
It is originally designed for video captioning task, and recently introduced into the task of moment localization with natural language, since these two tasks are reversible~\cite{chen2018temporally,zhang2019cross}.
Following the experimental setting in \cite{zhang2019cross}, we use val\_1 as validation set and val\_2 as testing set , which have $37,417$, $17,505$, and $17,031$ moment-sentence pairs for training, validation, and testing, respectively. Currently, this is the largest dataset in this task.

\textit{TACoS.}
It consists of 127 videos selected from the MPII Cooking Composite Activities video corpus~\cite{rohrbach2012script}, which contains different activities happened in kitchen room.
Regneri \textit{et al.} extends the sentence descriptions by crowd-sourcing.
A standard split~\cite{gao2017tall} consists of $9,790$, $4,436$, and $4,001$ moment-sentence pairs for training, validation and testing, respectively.

\subsection{Experimental Settings}

\subsubsection{Evaluation Metric.}
Following the setting as previous work~\cite{gao2017tall}, we evaluate our model by computing $Rank$ $n$@$m$. It is defined as the percentage of language queries having at least one correct moment retrieval in the top-$n$ retrieved moments. A retrieved moment is correct when its IoU with the ground truth moment is larger than $m$. 
There are specific settings of $n$ and $m$ for different datasets. Specifically,
we report the results as $n\in\{1, 5\}$ with $m\in\{0.5,0.7\}$ for Charades-STA dataset, $n\in\{1, 5\}$ with $m\in\{0.3,0.5,0.7\}$ for ActivityNet Captions dataset, and $n\in\{1, 5\}$ with $m\in\{0.1,0.3,0.5\}$ for TACoS dataset.

\subsubsection{Implementation Details.}
We use Adam~\cite{kingma2014adam} with learning rate of $1\small{\times}10^{-4}$
and batch size of $32$ for optimization.
A three layer LSTM is used for language encoding. The size of all hidden states (i.e. $d^S$, $d^V$ and $d^O$) in the model is set to $512$.
For a fair comparison, we adopt the same visual features as previous work~\cite{zhang2019man,zhang2019cross}, i.e., VGG feature~\cite{Simonyan15} for Charades, and C3D~\cite{tran2015learning} feature for ActivityNet Captions and TACoS.
the number of frames in a clip $T$ is set to $4$ for Charades-STA, and $16$ for ActivityNet Captions and TACoS.
On TACoS, the overlapping between neighboring clips is set to 0.8, while on Charades-STA and ActivityNet, the overlapping is set to 0, i.e. no overlapping.
The number of sampled clips $N$ is set to $16$ for Charades-STA, $64$ for ActivityNet Captions, and $128$ for TACoS.
Non maximum suppression (NMS) with a threshold of $0.5$ is applied during the inference.
For 2D-TAN network architecture, we adopt an $8$-layer convolution network with kernel size of $5$ for Charades-STA and TACoS (i.e. $L$=$8$ and $K$=$5$), and a $4$-layer convolution network with kernel size of $9$ for ActivityNet Captions (i.e. $L$=$4$ and $K$=$9$).
The scaling thresholds $t_{min}$ and $t_{max}$ are set to $0.5$ and $1.0$ for Charades-STA and ActivityNet Captions, and $0.3$ and $0.7$ for TACoS.

\subsection{Comparison to State-of-the-Art Methods}
We evaluate the proposed 2D-TAN approach on three benchmark datasets, and compare it with recently proposed state-of-the-art methods, including:
\begin{itemize}[leftmargin=0.38cm]
\item{
sliding window based methods: 
MCN~\cite{hendricks17iccv},  CTRL~\cite{gao2017tall},  ACRN~\cite{liu2018attentive},  ACL-K~\cite{Ge_2019_WACV} and  VAL~\cite{song2018val},}
\item{RNN-based methods: 
TGN~\cite{chen2018temporally} and CMIN~\cite{zhang2019cross},}
\item{
GCN-based method:
MAN~\cite{zhang2019man}, }
\item{
and others:
ROLE~\cite{liu2018crossmodal},  QSPN~\cite{xu2019multilevel},  SM-RL~\cite{wang2019language},  SLTA~\cite{jiang2019cross},  ABLR~\cite{yuan2019to},  SAP~\cite{chen2019semantic}, TripNet~\cite{Hahn2019tripping} and MCF~\cite{wu2018multi}.}
\end{itemize}
The results are summarized in Table~\ref{tab:Charades}--\ref{tab:TACoS}.

\begin{table}[t]
\small
	\begin{center}
		\begin{tabular}{|c|c|c|c|c|c|}
			\hline
			\multicolumn{2}{|c|}{\multirow{2}*{Method}} & \multicolumn{2}{c|}{$Rank1@$} & \multicolumn{2}{c|}{$Rank5@$} \\
			\cline{3-6}
			\multicolumn{2}{|c|}{} & $0.5$ & $0.7$ & $0.5$ & $0.7$ \\
			\hline
			\multicolumn{2}{|c|}{MCN}  & $17.46$ & $8.01$ & $48.22$ & $26.73$ \\
			\multicolumn{2}{|c|}{CTRL}  & $23.63$ & $8.89$ & $58.92$ & $29.52$ \\
			\multicolumn{2}{|c|}{ACRN}  & $20.26$ & $7.64$ & $71.99$ & $27.79$ \\
			\multicolumn{2}{|c|}{ROLE}  & $21.74$ & $7.82$ & $70.37$ & $30.06$ \\
			\multicolumn{2}{|c|}{VAL} & $23.12$ & $9.16$ & $61.26$ & $27.98$ \\
			\multicolumn{2}{|c|}{ACL-K}  & $30.48$ & $12.20$ & $64.84$ & $35.13$ \\
			\multicolumn{2}{|c|}{MAN}  & $\mathbf{41.24}$ & $\mathit{20.54}$ & $\mathbf{83.21}$ & $\mathit{51.85}$ \\
			\multicolumn{2}{|c|}{QSPN}  & $35.60$ & $15.80$ & $79.40$ & $45.40$ \\
			\multicolumn{2}{|c|}{SM-RL} & $24.36$ & $11.17$ & $61.25$ & $32.08$ \\
			\multicolumn{2}{|c|}{SLTA}  & $22.81$ & $8.25$ & $72.39$ & $31.46$ \\
			\multicolumn{2}{|c|}{ABLR}  & $24.36$ & $9.01$ & $-$ & $-$ \\
			\multicolumn{2}{|c|}{SAP} & $27.42$ & $13.36$ & $66.37$ & $38.15$ \\
			\multicolumn{2}{|c|}{TripNet} & $36.61$ & $14.50$ & $-$ & $-$  \\
			\hline
			\multirow{3}{*}{\rotatebox{90}{\textbf{2D-TAN}}}&  & &  & & \\[-1.5ex]
			& Pool  & $39.70$ & $\mathbf{23.31}$ & $\mathit{80.32}$ & $51.26$ \\[1.5ex]
			&Conv  & $\mathit{39.81}$ & ${23.25}$ & $79.33$ & $\mathbf{52.15}$ \\[1ex]
			\hline
		\end{tabular}
	\end{center}
	\caption{Performance comparison on Charades-STA. 
	Pool and Conv represent max-pooling and stacked convolution respectively, which indicates two different ways for moment feature extraction in our 2D-TAN.
	The values highlighted by \textbf{bold} and \textit{italic} fonts indicate the top-$2$ methods, respectively.
	The remaining tables follow the same notations.
	}
	\label{tab:Charades}
\end{table}

The results show that 2D-TAN performs among the best in various scenarios on all three benchmark datasets across different criteria. In all cases, 2D-TAN ranks the first or the second. It is worth noting that on TACoS dataset (see Table~\ref{tab:TACoS}), our 2D-TAN surpasses the state-of-the-arts, i.e. ACL-K 
and TGN
, by more than $5$ points and $14$ points in term of $Rank1@0.5$ and $Rank5@0.5$, respectively. Moreover, on the large-scale ActivityNet Captions dataset, 2D-TAN also outperforms the top ranked method CMIN with repect to $IoU@0.5$ and $0.7$. It validates that 2D-TAN is able to localize the moment boundary more precisely.

In more details, by comparing 2D-TAN with other related methods, we obtain serveral observations.
First, we compare 2D-TAN with previous sliding window based methods: MCN, CTRL, ACRN, ACL-K and VAL.
From the results in Table~\ref{tab:Charades}--\ref{tab:TACoS}, we observe that 
our 2D-TAN achieves superior results than sliding window methods. The reason is that independently matching the sentence with moment candidates ignores the temporal dependencies, and cannot distinguish the small differences between overlapped moments.
Differently, our proposed 2D-TAN models the dependencies between moment candiates by a 2D temporal map, and enables the network to perceive more context information from the adjacent moment candidates. Hence, it gains large improvements compared to sliding window based methods.

Moreover, we compare our approach with RNN-based methods, i.e. TGN and CMIN. 
Due to the involvement of context information during prediction, the RNN-based approaches perform better than the sliding window approaches, however, inferior to our proposed 2D-TAN method. 
RNN-based approaches implicitly update the context information through a recurrent memory module, while our 2D-TAN explicitly exploit the context information via a 2D temporal map. 
It further verifies the effectiveness of our model in high quality moment localization.

Last, we compare our method with graph convolutional netowrk (GCN) based method MAN~\cite{zhang2019man}, which achieves the state-of-the-art on Charades-STA. It utilizes a GCN to model the relations between moment pairs. Differently, our 2D-TAN models the temporal dependencies through a 2D convolution network.
From Table~\ref{tab:Charades}, we can see that 2D-TAN performs better at higher $IoU$@$0.7$, while slightly inferior to MAN at lower $IoU$@$0.5$.

\begin{table}[t]
\small
	\begin{center}
		\begin{tabular}{|P{0.1cm}|P{0.6cm}|P{0.7cm}|P{0.7cm}|P{0.7cm}|P{0.7cm}|P{0.7cm}|P{0.7cm}|}
			\hline
			\multicolumn{2}{|c|}{\multirow{2}*{Method}} & \multicolumn{3}{c|}{$Rank1@$} & \multicolumn{3}{c|}{$Rank5@$} \\
			\cline{3-8}
			\multicolumn{2}{|c|}{} & $0.3$ & $0.5$ & $0.7$ & $0.3$ & $0.5$ & $0.7$ \\
			\hline
			\multicolumn{2}{|c|}{MCN}  & $39.35$ & $21.36$ & $6.43$ & $68.12$ & $53.23$ & $29.70$  \\
			\multicolumn{2}{|c|}{CTRL}  & $47.43$ & $29.01$ & $10.34$ & $75.32$ & $59.17$ & $37.54$  \\
			\multicolumn{2}{|c|}{TGN}  & $43.81$ & $27.93$ & $-$ & $54.56$ & $44.20$ & $-$  \\
			\multicolumn{2}{|c|}{ACRN}  & $49.70$ & $31.67$ & $11.25$ & $76.50$ & $60.34$ & $38.57$  \\
			\multicolumn{2}{|c|}{CMIN}  & $\mathbf{63.61}$ & $\mathit{43.40}$ & $\mathit{23.88}$ & $\mathit{80.54}$ & $\mathit{67.95}$ & $\mathit{50.73}$  \\
			\multicolumn{2}{|c|}{QSPN}  & $52.13$ & $33.26$ & $13.43$ & $77.72$ & $62.39$ & $40.78$  \\
			\multicolumn{2}{|c|}{ABLR}  & $55.67$ & $36.79$ & $-$ & $-$ & $-$ & $-$  \\
			\multicolumn{2}{|c|}{TripNet} & $48.42$ & $32.19$ & $13.93$ & $-$ & $-$ & $-$  \\
			\hline
			\multirow{3}{*}{\rotatebox{90}{\textbf{2D-TAN}}}& & & & & & & \\[-1.5ex]
			& Pool & $\mathit{59.45}$ & $\mathbf{44.51}$ & ${26.54}$ & ${85.53}$ & $\mathbf{77.13}$ & ${61.96}$  \\[1.5ex]
			& Conv & $58.75$ & ${44.05}$ & $\mathbf{27.38}$ & $\mathbf{85.65}$ & ${76.65}$ & $\mathbf{62.26}$ \\[1ex]
			\hline
		\end{tabular}
	\end{center}
	\caption{Performance comparison on ActivityNet Captions.
}
	\label{tab:ActivityNet Captions}
\end{table}

\begin{table}[t]
\small
	\begin{center}
		\begin{tabular}{|P{0.1cm}|P{0.6cm}|P{0.7cm}|P{0.7cm}|P{0.7cm}|P{0.7cm}|P{0.7cm}|P{0.7cm}|}
			\hline
			\multicolumn{2}{|c|}{\multirow{2}*{Method}} & \multicolumn{3}{c|}{$Rank1@$} & \multicolumn{3}{c|}{$Rank5@$} \\
			\cline{3-8}
			\multicolumn{2}{|c|}{} & $0.1$ & $0.3$ & $0.5$ & $0.1$ & $0.3$ & $0.5$ \\
			\hline
			\multicolumn{2}{|c|}{MCN}  & $14.42$ & $-$ & $5.58$ & $37.35$ & $-$ & $10.33$  \\
			\multicolumn{2}{|c|}{CTRL}  & $24.32$ & $18.32$ & $13.30$ & $48.73$ & $ 36.69$ & $25.42$  \\
			\multicolumn{2}{|c|}{MCF}  & $25.84$ & $18.64$ & $12.53$ & $52.96$ & $37.13$ & $24.73$  \\
			\multicolumn{2}{|c|}{TGN}  & $41.87$ & $21.77$ & $18.9$ & $53.40$ & $39.06$ & $\mathit{31.02}$  \\
			\multicolumn{2}{|c|}{ACRN}  & $24.22$ & $19.52$ & $14.62$ & $47.42$ & $34.97$ & $24.88$  \\
			\multicolumn{2}{|c|}{ROLE}  & $20.37$ & $15.38$ & $9.94$ & $45.45$ & $31.17$ & $20.13$  \\
			\multicolumn{2}{|c|}{VAL} & $25.74$ & $19.76$ & $14.74$ & $51.87$ & $38.55$ & $26.52$  \\
			\multicolumn{2}{|c|}{ACL-K}  & $31.64$ & $24.17$ & $\mathit{20.01}$ & $57.85$ & $\mathit{42.15}$ & ${30.66}$  \\
			\multicolumn{2}{|c|}{CMIN}  & $32.48$ & $\mathit{24.64}$ & $18.05$ & $\mathit{62.13}$ & $38.46$ & $27.02$  \\
			\multicolumn{2}{|c|}{QSPN}  & $25.31$ & $20.15$ & $15.23$ & $53.21$ & $36.72$ & $25.30$  \\
			\multicolumn{2}{|c|}{SM-RL} & $26.51$ & $20.25$ & $15.95$ & $50.01$ & $38.47$ & $27.84$  \\
			\multicolumn{2}{|c|}{SLTA}  & $23.13$ & $17.07$ & $11.92$ & $46.52$ & $32.90$ & $20.86$  \\
			\multicolumn{2}{|c|}{ABLR}  & $\mathit{34.70}$ & $19.50$ & $9.40$ & $-$ & $-$ & $-$  \\
			\multicolumn{2}{|c|}{SAP} & $31.15$ & $-$ & $18.24$ & $53.51$ & $-$ & $28.11$  \\
			\multicolumn{2}{|c|}{TripNet} & $-$ & $23.95$ & $19.17$ & $-$ & $-$ & $-$ \\
			\hline
			\multirow{3}{*}{\rotatebox{90}{\textbf{2D-TAN}}}& & & & &  & & \\[-1.5ex]
			&Pool & $\mathbf{47.59}$ & $\mathbf{37.29}$ & $\mathbf{25.32}$ & ${70.31}$ & $\mathbf{57.81}$ & $\mathbf{45.04}$ \\[1.5ex]
			&Conv & $46.44$ & $35.22$ & ${25.19}$ & $\mathbf{74.43}$ & ${56.94}$ & $44.21$ \\[1ex]
			\hline
		\end{tabular}
	\end{center}
	\caption{Performance comparison on TACoS. 
}
	\label{tab:TACoS}
\end{table}

\subsection{Ablation Study}

\begin{table*}[t]
\small
	\begin{center}
		\begin{tabular}{|c|c|c|c|c|c|c|c|c|c|c|c|c|}
			\hline
			\multirow{2}*{Row\#} & \multicolumn{2}{c|}{\multirow{2}*{Method}} &\multirow{2}*{$N$} & \multicolumn{2}{c|}{2D-TAN}  & \multicolumn{3}{c|}{$Rank1@$} & \multicolumn{3}{c|}{$Rank5@$}\\
			\cline{5-12}
			& \multicolumn{2}{c|}{} & & Kernel & Layer & $0.3$ & $0.5$ & $0.7$ & $0.3$ & $0.5$ & $0.7$\\
			\hline
			$1$ & \multicolumn{2}{c|}{Upper Bound} & $16$ & $-$ & $-$ & $97.16$ & $93.58$ & $89.14$ & $97.16$ & $93.58$ & $89.14$\\
			$2$ & \multicolumn{2}{c|}{Upper Bound} & $32$ & $-$ & $-$ & $99.10$ & $96.88$ & $94.38$ & $99.10$ & $96.88$ & $94.38$\\
			$3$ & \multicolumn{2}{c|}{Upper Bound} & $64$ & $-$ & $-$ & $99.84$ & $98.94$ & $97.34$ & $99.84$ & $98.94$ & $97.34$\\
			\hline
			$4$ & & Enum & $16$ & $9$ & $4$ & $58.82$ & $42.45$ & $23.93$ & $85.07$ & $75.99$ & $57.79$ \\
			$5$ & & Enum & $32$ & $9$ & $4$ & $58.26$ & $43.18$ & $25.47$ & $84.82$ & $75.45$ & $59.66$ \\
			$6$ & & Enum & $64$ & $9$ & $4$ & $58.15$ & $42.80$ & $25.76$ & $84.53$ & $75.39$ & $60.18$ \\
			$7$ & & Enum & $64$ & $1$ & $1$ & $45.90$ & $26.20$ & $14.27$ & $70.72$ & $56.14$ & $37.13$ \\
			$8$ & \multirow{3}{*}{\rotatebox{90}{\textbf{2D-TAN}}} & Enum & $64$ & $5$ & $1$ & $54.78$ & $35.27$ & $18.81$ & $81.80$ & $69.76$ & $50.68$ \\
			$9$ & & Enum & $64$ & $5$ & $4$ & $58.20$ & $40.45$ & $23.25$ & $83.76$ & $73.97$ & $57.46$ \\
			$10$ & & Enum & $64$ & $9$ & $4$ & $58.15$ & $42.80$ & $25.76$ & $84.53$ & $75.39$ & $60.18$ \\
			$11$ & & Pool &$64$ & $9$ & $4$ & $59.45$ & $44.51$ & $26.54$ & $85.53$ & $77.13$ & $61.96$ \\
			$12$ & & Pool &$64$ & $5$ & $8$ & $57.86$ & $41.68$ & $25.13$ & $85.26$ & $75.74$ & $58.90$ \\
			$13$ & & Pool &$64$ & $17$ & $2$ & $58.19$ & $43.09$ & $26.09$ & $84.22$ & $75.16$ & $60.02$ \\
			$14$ & & Conv &$64$ & $9$ & $4$  & $58.75$ & ${44.05}$ & ${27.38}$ & ${85.65}$ & ${76.65}$ & ${62.26}$\\
			\hline
		    \hline
			$15$ & \multicolumn{2}{c|}{CTRL} & $-$ & $-$ & $-$ & $47.43$ & $29.01$ & $10.34$ & $75.32$ & $59.17$ & $37.54$  \\
			$16$ & \multicolumn{2}{c|}{CMIN} & $200$ & $-$ & $-$ & $63.61$ & $43.40$ & $23.88$ & $80.54$ & $67.95$ & $50.73$\\
			\hline
		\end{tabular}
	\end{center}
	\caption{Ablation Study.
	$N$ is the number of sampled clips.
	Row $1-3$ show the upper bound of an ideal model under different $N$.
	Row $4-6$ demonstrate how our model perform under different $N$.
	Row $6-13$ compare the performance under different kernel and layer settings.
	Row $14$ show the performance using moment features  extracted by stacked convolution.
	Row $15-16$ are two previous methods for comparison.
	}
	\label{tab:ablation_study}
\end{table*}

In this section, we evaluate the effects of different factors in our proposed 2D-TAN.
The experiments are conducted on the ActivityNet Captions dataset, as shown in Table~\ref{tab:ablation_study}.

\textit{Number of Moment Candidates. } 
The number of moment candidates is a vital factor in moment localization models.
We first tune this factor in our 2D-TAN approach, and show its impacts on final performance. Then, we compare different approaches with respect to this factor.

We vary the number of sampled clips $N$ from $16$ to $64$ in our 2D-TAN approach. The results are shown in Table~\ref{tab:ablation_study} (Row $4-6$). We observe that, increasing $N$ from $16$ to $64$ brings  improvements ($57.79$ $v.s$ $59.66$ $v.s$ $60.18$ in $Rank5@0.7$). This observation is also consistent with the theoretical upper bound, as listed in Table~\ref{tab:ablation_study} (Row $1-3$). Here, the upper bound represents the performance of an ideal model that can provide a correct prediction on all the sampled video clips. The upper bound is smaller than $100\%$ since that the sampling of video clips introduces errors.

Furthermore, we compare the number of moment candidates with the previous state-of-the-art method CMIN.
Row $16$ in Table~\ref{tab:ablation_study} shows that CMIN use $1400$ ($N$=$200$) moment candidates. However, our 2D-TAN only uses $136$ ($N$=$16$) candidates, and achieves comparable results to CMIN (Row $4$ \emph{v.s} $16$ ).
Moreover, with larger number of moment candidates ($N$=$64$) and stacked convolution layers for moment representations, the performance of our method can be further boosted, as listed in Row $14$.
Noted that the number of moment candidates in Row $14$ is $1200$, which is still smaller than the ones used in CMIN.
This comparison validates that our 2D-TAN gains improvement from the context modeling, rather than the dense sampling of moment candidates.

\textit{Receptive Field Size. } We vary the depth and kernel size of convolution layers in our 2D-TAN to study the impact of receiptive field size.  
The results in terms of different kernel sizes and layer depths are reported in Table~\ref{tab:ablation_study} Row $7-9$. 
We observe that the performance increases significantly as the receptive field enlarges.
However, it becomes saturated when it is large enough, as listed in Row $6$.
Moreover, if the receptive field size is fixed, changing the depth of layers and kernel sizes has limited impacts on final performance, as shown in Row $11$-$13$. 
This verifies the importance of receiptive field size in our 2D-TAN model. Large receiptive field is able to model temporal dependencies, resulting in performance improvements.
If we set the kernel size to $1$ (Row $7$), the 2D-TAN model is equivalent to treat each moment independently. In this case, it achieves similar performance with CTRL method (Row $15$), which also treats each moment individually. 
This phenomenon further proves our hypothesis that modeling the moment candidates as a whole enables the network to distinguish similar moments.

\textit{Sparse Sampling \emph{v.s.} Enumeration.}
We further compare the effectiveness of our sparse sampling strategy with the dense enumeration for moment candidate selection. The results are reported in Table~\ref{tab:ablation_study} (Row $10$-$11$). It is observed that these two strategies achieve similar performance. The underlying reason is that the designed sparse sampling removes nearly $50\%$ redundant moment candidates. Thus, it reduces the computation cost without performance decrease.

 \textit{Stacked Convolution \emph{v.s.} Max-Pooling.} 
Stacked convolution and pooling have been applied for extracting moment features in previous works~\cite{hendricks17iccv,zhang2019man}.
We compare their performance on three datasets, as  shown in Table~\ref{tab:Charades}-\ref{tab:TACoS} (2D-TAN: Pool \emph{v.s.} Conv).
It is observed that stacked convolution (Conv) performs better than max-pooling (Pool) on ActivityNet Captions, while comparable on Charades-STA and TACoS. We recommend to adopt the max-pooling operation, since it is fast in calculation, while does not contain any parameters.

\section{Conclusion}

In this paper, we study the problem of moment localization with natural language, and propose a novel
2D Temporal Adjacent Networks(2D-TAN) method.
The core idea is to retrieve a moment on a two-dimensional temporal map, which considers adjacent moment candidates as the temporal context. 2D-TAN is capable of encoding adjacent temporal relation, while learning discriminative feature for matching video moments with referring expressions.
Our model is simple in design and achieves  competitive performance in comparison with the state-of-the-art methods on three benchmark datasets.
In the future, we would like to extend our model to other temporal localization tasks, such as temporal action localization, video re-localization, etc. 

\section{Acknowledgement}
We thank the support of NSF awards IIS-1704337, IIS-1722847, IIS-1813709, and the generous gift from our corporate sponsors.

\bibliography{reference}
\bibliographystyle{aaai}
\end{document}